\pdfoutput=1

\documentclass[11pt,letterpaper]{article}

\usepackage{emnlp2017}
\usepackage{times}
\usepackage{latexsym}
\usepackage{amssymb}
\usepackage{graphicx}
\usepackage{amsmath}
\usepackage{url}
\usepackage{booktabs}
\usepackage{array}
\usepackage{multirow}
\usepackage{xcolor}
\usepackage[T1]{fontenc}

\newcommand{\MC}[3]{\multicolumn{#1}{#2}{#3}}
\newcommand{\MR}[3]{\multirow{#1}{#2}{#3}}

\newcommand{\tbf}[1]{\textbf{#1}}
\newcommand{\tit}[1]{\textit{#1}}
\newcommand{\tul}[1]{\underline{#1}}
\newcommand{\tm}[1]{\texttt{#1}}

\newcommand{\ra}{$\rightarrow$}
\newcommand{\htoo}{\textit{h2o}}

\emnlpfinalcopy

\def\emnlppaperid{92}

\title{LIUM Machine Translation Systems for WMT17 News Translation Task}

\author{Mercedes Garc\'ia-Mart\'inez, Ozan Caglayan$^\dagger$, Walid Aransa \\
       \bf Adrien Bardet, Fethi Bougares, Lo\"ic Barrault \\
        LIUM, University of Le Mans \\
      $^\dagger$\tm{ozancag@gmail.com} \\
      \tm{FirstName.LastName@univ-lemans.fr}}

\begin{document}
\maketitle

\begin{abstract}
This paper describes LIUM submissions to WMT17 News Translation Task
for English$\leftrightarrow$German, English$\leftrightarrow$Turkish,
English$\rightarrow$Czech and English$\rightarrow$Latvian language pairs.
We train BPE-based attentive Neural Machine Translation systems with and without
factored outputs using the open source \tit{nmtpy} framework.
Competitive scores were obtained by ensembling various systems and
exploiting the availability of target monolingual corpora for back-translation.
The impact of back-translation quantity and quality is also analyzed
for English\ra Turkish where our post-deadline submission surpassed the best entry by +1.6 BLEU.
\end{abstract}

\section{Introduction}
This paper describes LIUM Neural Machine Translation (NMT) submissions to WMT17 News Translation Task
for English$\leftrightarrow$German, English$\leftrightarrow$Turkish,
English$\rightarrow$Czech and English$\rightarrow$Latvian language pairs.
We experimented with and without back-translation data for English$\leftrightarrow$German
and English$\leftrightarrow$Turkish which are respectively described in Sections~\ref{sec:en_tr} and~\ref{sec:ende}.
For the latter pair, we also present an analysis about the impact of back-translation quality and quantity as well as two architectural
ablations regarding the initialization and the output of recurrent decoder (Section~\ref{sec:en_tr}).

Experiments for English$\rightarrow$Czech and English$\rightarrow$Latvian are performed using Factored NMT (FNMT) \cite{Garcia16iwslt} systems.
FNMT is an extension of NMT which aims at simultaneously predicting the canonical form of a word
and its morphological information needed to generate the final surface form. The details and results are presented in section~\ref{fnmt}.
All submitted systems\footnote{Backtranslations and other data can be found at \url{http://github.com/lium-lst/wmt17-newstask}}
are trained using the open source \tm{nmtpy}\footnote{\url{http://github.com/lium-lst/nmtpy}} framework \cite{nmtpy}.

\section{Baseline NMT}
\label{sec:nmt}
Our baseline NMT is an attentive encoder-decoder \cite{Bahdanau2014} implementation.
A bi-directional Gated Recurrent Unit (GRU) \cite{Chung2014} encoder is used to compute source sentence annotation vectors.
We equipped the encoder with layer normalization \cite{ba2016layer}, a technique which adaptively normalizes the incoming activations
of each hidden unit with a learnable gain and bias, after empirically observing that it improves both convergence speed and translation performance.

A conditional GRU (CGRU) \cite{cgru,nematus} decoder with attention mechanism
is used to generate a probability distribution over target tokens for each decoding step $t$.
The hidden state of the CGRU is initialized using a non-linear transformation of the average encoder state
produced by the encoder.
Following \newcite{inan2016tying, press2016using}, the feedback embeddings (input to the decoder) and
the output embeddings are \tbf{tied} to enforce learning a single target representation
and decrease the number of total parameters by target vocabulary size $\times$ embedding size.

We used Adam \cite{kingma2014adam} as the optimizer with a learning rate of $4e\mathrm{-}4$.
Weights are initialized with Xavier scheme \cite{glorotxavier} and the total gradient norm is
clipped to 5 \cite{pascanu2013difficulty}.
When stated, three dropouts \cite{srivastava2014dropout} are applied after
source embeddings, encoder hidden states and pre-softmax activations respectively.
The training is early stopped if validation set BLEU~\cite{bleu2002} does not improve for
a given number of consecutive validations. A beam size of \tbf{12} is used for beam-search decoding.
Other hyper-parameters including layer dimensions and dropout probabilities are detailed
for each language pair in relevant sections.

\section{English$\leftrightarrow$Turkish}
\label{sec:en_tr}
\newcommand{\ph}[1]{\phantom{#1}}
\subsection{Training}
\label{sec:entr:train}
We use SETIMES2 which consists of 207K parallel sentences
for training, newsdev2016 for early-stopping, and newstest2016 for model selection (internal test).
All sentences are normalized and tokenized using \tit{normalize-punctuation}
and \tit{tokenizer} \footnote{The tokenizer is slightly modified
to fix handling of apostrophe splitting in Turkish.} from Moses \cite{Moses:2007:acl}.
Training sentences that have less than 3 and more than 50 words are filtered out and
a joint Byte Pair Encoding (BPE) model \cite{sennrich2015neural} with 16K merge operations is learned
on train+newsdev2016. The resulting training set has 200K sentences and 5.5M tokens (Table~\ref{tbl:en_tr_data})
where $\sim$63\% and $\sim$50\% of English and Turkish vocabularies is composed of a common set of tokens.

\begin{table}[ht]
\centering
\resizebox{.8\columnwidth}{!}{%
\begin{tabular}{rl}
\toprule
\tbf{Language} & \tbf{\# BPE Tokens} \\ \midrule
English        & 10041 = 6285 Common + 3756 En \\
Turkish        & 12433 = 6285 Common + 6148 Tr \\ \midrule
Combined       & 16189 \\ \bottomrule
\end{tabular}
}
\caption{Sub-word statistics for English, Turkish and Combined vocabularies.}
\label{tbl:en_tr_data}
\end{table}

All models use \tbf{200}-dimensional embeddings and GRU layers with \tbf{500} hidden units.
The dropout probability $P_{drop}$ is used for all 3 dropouts and set to 0.2 and 0.3 for EN\ra TR and TR\ra EN respectively.
The validation BLEU is computed after each $\sim$1/4 epoch and the training stops if no
improvement is achieved after 20 consecutive validations.

\paragraph{Data Augmentation}
Due to the low-resource characteristic of EN$\leftrightarrow$TR,
additional training data has been constructed using back-translations (BT) \citep{backtrans}
where target-side monolingual data is translated to source language to
form a Source\ra Target synthetic corpus. newscrawl2016 (1.7M sentences) and newscrawl2014 (3.1M sentences)
are used as monolingual data for Turkish and English respectively.
Although we kept the amount of synthetic data around $\sim$150K sentences for submitted systems
to preserve \tit{original-to-synthetic} ratio,
we present an analysis about the impact of synthetic data quantity/quality as a follow-up
study in Section~\ref{sec:followup_en_tr}. All back-translations are produced using the NMT
systems described in this study.

\paragraph{3-way Tying (3WT)}
In addition to tying feedback and output embeddings (Section~\ref{sec:nmt}),
we experiment with 3-way tying (3WT) \cite{press2016using} only for \tbf{EN\ra TR} where we
use the \tbf{same} embeddings for source, feedback and output embeddings.
A \tit{combined} vocabulary of $\sim$16K tokens (Table~\ref{tbl:en_tr_data}) is then used
to form a bilingual representation space.

\paragraph{Init-0 Decoder}
The attention mechanism \cite{Bahdanau2014}
introduces a time-dependent context vector (weighted sum of encoder states)
as an auxiliary input to the decoder allowing implicit encoder-to-decoder connection
through which the error back-propagates towards source embeddings.
Although this makes it unnecessary to initialize the decoder, the first hidden
state of the decoder is generally derived from the last \cite{Bahdanau2014} or average
encoder state \cite{nematus} in common practice.
To understand the impact of this, we train additional \tbf{Init-0} EN\ra TR systems
where the decoder is initialized with an all-zero vector instead of average encoder state.

\subsection{Submitted Systems}
Each system is trained twice with different seeds and the one with better newstest2016 BLEU
is kept when reporting single systems. Ensembles by default use the best early-stop
checkpoints of both seeds unless otherwise stated. Results for \tit{both} directions are presented in Table~\ref{tbl:en_tr_res}.

\tbf{TR\ra EN} baseline (E1) achieves 14.2 BLEU on
newstest2017. The (E2) system trained with additional
150K BT data surpasses the baseline by $\sim$2 BLEU on newstest2017.
The EN\ra TR system used for BT is a single (T5) system which is itself a BT-enhanced NMT.
A contrastive system (E3) with less dropout ($P_{drop}=0.2$)
is used for our final submission which is an ensemble of 4 systems (2 runs of E2 + 2 runs of E3).
In overall, an improvement of $\sim$3.7 BLEU over the baseline system is achieved by
making use of a small quantity of BT data and ensembling.

\tbf{EN\ra TR} baseline (T1) achieves 11.1 BLEU on newstest2017
(Table~\ref{tbl:en_tr_res}). (T2) which is augmented with 150K synthetic data,
improves over (T1) by 2.5 BLEU. It can be seen that once 3-way tying (3WT)
is enabled, a consistent improvement of up to 0.6 BLEU is obtained on newstest2017.
We conjecture that 3WT is beneficiary (especially in a low-resource regime)
when the intersection of vocabularies is a large set since the embedding
of a common token will now receive as many updates as its occurrence count in both sides of the corpus.
On the other hand, the initialization method of the decoder does not seem to incur
a significant change in BLEU. Finally, using an ensemble of 4 3WT-150K-BT systems with
different decoder initializations (2xT5 + 2xT6), an overall improvement of 4.9 BLEU is obtained over (T1).
As a side note, 3WT reduces the number of parameters by $\sim$10\% (12M\ra 10.8M).

\begin{table}[htbp!]
\centering
\resizebox{1.\columnwidth}{!}{%
\begin{tabular}{lccc}
\toprule
\tbf{System}                        & \tbf{3WT}     & \tbf{nt2016} & \tbf{nt2017}   \\ \midrule
\MC{4}{c}{\tbf{TR\ra EN} ($P_{drop}=0.3$)} \\ \midrule
(E1) Baseline (200K)                & $\times$   & 14.2            & 14.2            \\
(E2) E1 + 150K-BT                   & $\times$   & 16.6            & \tul{16.1}      \\
(E3) E1 + 150K-BT ($P_{drop}=0.2$)  & $\times$   & 16.4            & 16.3            \\ \midrule
Ensemble (2xE2 + 2xE3)              & $\times$   & 18.1            & \tbf{17.9}      \\ \midrule \midrule
\MC{4}{c}{\tbf{EN\ra TR} ($P_{drop}=0.2$)} \\ \midrule
(T1) Baseline (200K)                & $\times$      & 10.9           & 11.1             \\            
(T2) T1 + 150K-BT                   & $\times$      & 12.7           & 13.6             \\            
(T3) T1 + 150K-BT + Init0           & $\times$      & 12.8           & 13.5             \\            
(T4) Baseline (200K)                & $\checkmark$  & 11.5           & 11.6             \\   
(T5) T4 + 150K-BT                   & $\checkmark$  & 13.4           & \tul{14.2}       \\            
(T6) T4 + 150K-BT + Init0           & $\checkmark$  & 13.3           & 14.0             \\ \midrule   
Ensemble (2xT5 + 2xT6)              & $\checkmark$  & 14.7           & \tbf{16.0}       \\ \bottomrule
\end{tabular}}
\caption{EN$\leftrightarrow$TR: \tul{Underlined} and \tbf{bold} scores represent contrastive and primary submissions respectively.}
\label{tbl:en_tr_res}
\end{table}

\subsection{Follow-up Work}
\label{sec:followup_en_tr}
We dissect the output layer of CGRU NMT \cite{nematus}
which is conditioned (Equation~\ref{eq:cond1}) on the hidden state $h_t$ of the decoder,
the feedback embedding $y_{t-1}$ and the weighted context vector $c_t$. We experiment with
a \tit{simple output} (Equation~\ref{eq:cond2}) which depends only on $h_t$ similar to \newcite{Sutskever2014}.
The target probability distribution is computed (Equation~\ref{eq:softmax}) using softmax on top of this output transformed with $W_o$.

\begin{align}
\label{eq:cond1}
o_t &= \tanh (\mathbf{W_h}h_t + y_{t-1} + \mathbf{W_c}c_t)\\
\label{eq:cond2}
o_t &= \tanh(\mathbf{W_h}h_t) \\
P(y_t) &= \text{softmax}(\mathbf{W_o} o_t) \label{eq:softmax}
\end{align}

\begin{table}[htbp!]
\centering
\resizebox{1.\columnwidth}{!}{%
\begin{tabular}{lccccc}
\toprule
\MR{2}{*}{\tbf{System}} & \MR{2}{*}{\tbf{\# Sents}} & \MC{2}{c}{\tbf{nt2016}} & \MC{2}{c}{\tbf{nt2017}} \\
&                & Single       & Ens            & Single          & Ens  \\ \cmidrule(lr){1-2} \cmidrule(lr){3-4} \cmidrule(lr){5-6}
(B0) Only SETIMES2         & 200K & 11.5       & 12.8       & 11.6             & 13.0              \\
(B1) Only 1.0M-BT-E1       & 1.0M & 13.6       & 14.5       & 14.8             & 16.3 \\ \midrule
(B2) B0 + 150K-BT-E1       & 350K & 13.2       & 14.2       & 14.3             & 15.4              \\
(B3) \ph{B0 + 150K-}BT-E2  &      & 13.4       & 14.1       & 14.2             & 14.9              \\ \midrule
(B4) B0 + 690K-BT-E1       & 890K & 14.8       & 15.4       & 15.9             & 17.1              \\
(B5) \ph{B0 + 690K-}BT-E2  &      & 14.7       & 15.6       & 16.1             & 16.9              \\ \midrule
(B6) B0 + 1.0M-BT-E1       & 1.2M & \tbf{14.9} & \tbf{15.6} & \tbf{16.2}       & \tbf{17.5} \\
(B7) \ph{B0 + 1.0M-}BT-E2  &      & 14.9       & 15.5       & 16.0             & 17.0              \\ \midrule
(B8) B0 + 1.7M-BT-E1       & 1.9M & 14.7       & 15.4       & 16.4             & 17.1              \\
(B9) \ph{B0 + 1.7M-}BT-E2  &      & 14.8       & 15.7       & 16.1             & 16.7              \\
\bottomrule
\end{tabular}}
\caption{Impact of back-translation quantity and quality for EN\ra TR: all systems are 3WT, (B0) is the same as (T4) from Table~\ref{tbl:en_tr_res}.}
\label{tbl:btquality}
\end{table}

As a second follow-up experiment, we analyse the impact of BT data quantity and quality
on final performance. Four training sets are constructed by taking the original 200K
training set and gradually growing it with BT data of size 150K, 690K, 1.0M and 1.7M (all-BT)
sentences respectively. The source side of the monolingual Turkish data used to create the synthetic corpus
are translated to English using two different TR\ra EN systems namely (E1) and (E2) where the latter is better
than former on newstest2016 by 2.4 BLEU (Table~\ref{tbl:en_tr_res}).

The results are presented in Table~\ref{tbl:btquality} and~\ref{tbl:finalres_en_tr}.
First, (B1) trained with \tit{only} synthetic data turns out to be superior than the baseline (B0) by 3.2 BLEU.
The ensemble of (B1) even surpasses our primary submission.
Although this may indicate the impact of training set size for NMT where
a large corpus with synthetic source sentences
leads to better performance than a human-translated but small corpus,
a detailed analysis would be necessary to reveal other possible reasons.

Second, it is evident that increasing the amount of BT data is beneficial
regardless of \tit{original-to-synthetic} ratio: the system (B6)
achieves +4.6 BLEU compared to (B0) on newstest2017 (11.6\ra16.2). The single (B6) is even slightly better than our
ensemble submission (Table~\ref{tbl:finalres_en_tr}). The +2.4 BLEU gap between back-translators E1 and E2 does not seem to affect
final performance where both groups achieve more or less the same scores.

Finally, the \tit{Simple Output} seems to perform slightly better than the original output formulation. In fact, our final
\tit{post-deadline} submission which surpasses the winning UEDIN system\footnote{\url{http://matrix.statmt.org}}
by 1.6 BLEU (Table~\ref{tbl:finalres_en_tr}) is an ensemble of four (B6) systems two of them being \tit{SimpleOut}.
Conditioning the target distribution over the weighted context vector $c_t$
creates an auxiliary gradient flow from the cross-entropy loss to the encoder
by skipping the decoder. We conjecture that conditioning only over the decoder's
hidden state $h_t$ forces the network (especially the decoder) to better
learn the target distribution. Same gradient flow also happens for feedback
embeddings in the original formulation (Equation~\ref{eq:cond1}).

\begin{table}[ht]
\centering
\resizebox{.95\columnwidth}{!}{%
\begin{tabular}{lcc}
\toprule
\tbf{System}                              & \tbf{Single}    & \tbf{Ens}       \\ \midrule
LIUM                                      & -               & 16.0      \\
UEDIN                                     & -               & \tbf{16.5}\\ \midrule \midrule
(B1) Only BT                              & 14.8            & 16.3      \\ \midrule
(B6) SETIMES2 + BT                        & 16.2            & 17.5      \\
(B6) + \tit{SimpleOut}                    & 16.6            & 17.6      \\ \midrule
Ensemble (2xB6 + 2xB6-\tit{SimpleOut})    & -               & \tbf{18.1}\\
\bottomrule
\end{tabular}}
\caption{Summary of follow-up results for EN\ra TR newstest2017: UEDIN is the best WMT17 matrix entry
before deadline while LIUM is our primary submission (Table~\ref{tbl:en_tr_res}).}
\label{tbl:finalres_en_tr}
\end{table}


\section{English$\leftrightarrow$German}
\label{sec:ende}
We train two types of model: first is trained with only parallel data provided by WMT17 (5.6M sentences),
the second uses the concatenation (9.3M sentences) of the provided parallel data and UEDIN WMT16
back-translation corpus \footnote{\url{http://data.statmt.org/rsennrich/wmt16_backtranslations}}.
Prior to training, all sentences are normalized, tokenized and truecased
using \emph{normalize-punctuation}, \emph{tokenizer} and \emph{truecaser} from Moses \cite{Moses:2007:acl}.
Training sentences with less than 2 and more than 100 units are filtered out.
A joint Byte Pair Encoding (BPE) model \cite{sennrich2015neural} with 50K merge operations is learned on the \tit{training data}.
This results in a vocabulary of 50K and 53K tokens for English and German respectively.

The training is stopped if no improvement is observed during 30 consecutive validations
on \emph{newstest2015}. Final systems are selected based on \emph{newstest2016} BLEU.

\subsection{Submitted Systems}
\paragraph{EN\ra DE} The baseline which is an NMT with \tbf{256}-dimensional embeddings and \tbf{512}-units GRU layers,
obtained 23.26 BLEU on newstest2017 (Table~\ref{tbl:ende_results}).
The addition of BT data improved this baseline by 1.7 BLEU (23.26\ra 24.94). Our primary
submission which achieved 26.60 BLEU is an ensemble of 4 systems: 2 best checkpoints of an NMT and 2 best checkpoints of an NMT
with 0-initialized decoder (See section~\ref{sec:entr:train}).

\paragraph{DE\ra EN} Our primary DE\ra EN system (Table~\ref{tbl:ende_results}) is an ensemble without back-translation (No-BT) of two NMT systems
with different dimensions: 256-512 and 384-640 for embeddings and GRU hidden units respectively.
Our post-deadline submission which is an ensemble with back-translation (BT) improved over our primary
system by +4.5 BLEU and obtained 33.9 BLEU on newstest2017. This ensemble consists
of 6 different systems (by varying the seed and the embedding and the GRU hidden unit size) trained with WMT17 and back-translation data.

\begin{table}[ht]
\centering
\resizebox{1.\columnwidth}{!}{%
\begin{tabular}{@{}rccc@{}}
\toprule
\tbf{System}                        & \tbf{\# Params} & \tbf{nt2016} & \tbf{nt2017}   \\ \midrule
EN$\rightarrow$DE Baseline          & 35.0M           & 29.11          & 23.26            \\
+ synthetic                         &                 & 31.08          & 24.94            \\
\ph{+ }primary ensemble             &                 & 33.89          & \tbf{26.60}      \\ \midrule \midrule
DE$\rightarrow$EN Baseline          & 52.9M           & 33.13          & 29.42            \\
primary ensemble (No-BT)            &                 & 33.63          & \tit{30.10}            \\ \midrule
+ synthetic                         &                 & 37.36          & 32.20            \\
post-deadline ensemble (BT)         &                 & \tbf{39.07}    & \tbf{33.90}      \\

\bottomrule
\end{tabular}}
\caption{BLEU scores computed with \textit{mteval-v13a.pl} for EN$\leftrightarrow$DE systems on newstest2016 and newstest2017.}
\label{tbl:ende_results}
\end{table}

\section{English$\rightarrow$\{Czech,Latvian\}}
\label{fnmt}
The language pairs English$\rightarrow$Czech and English$\rightarrow$Latvian are translated using a Factored NMT (FNMT) system where two symbols are generated at the same time.
The FNMT systems are compared to a baseline NMT system similar to the one described in Section~\ref{sec:nmt}.

\subsection{Factored NMT systems}
The FNMT system~\cite{Garcia16iwslt} is an extension of NMT where the lemma and the Part of Speech (PoS) tags of a word (i.e. factors) are produced at the output instead of its surface form.
The two output symbols are then combined to generate the word using external linguistic resources.
The low frequency words in the training set can benefit from sharing the same lemma with other high frequency words, and also from sharing the factors with other words having the same factors.
The lemma and its factors can sometimes generate new surface words which are unseen in the training data.
The vocabulary of the target language contains only lemmas and PoS tags but the total number of surface words that can be generated (i.e. virtual vocabulary) is larger because of the external linguistic resources that are used.
This allows the system to correctly generate words which are considered unknown words in word-based NMT systems.

We experimented with two types of FNMT systems which have a second output in contrast to baseline NMT.
The first one contains a single hidden to output layer ($\htoo$) which is then used by two separate softmaxes while
the second one contains two separate $\htoo$ layers each specialized for a particular output.
The lemma and factor sequences generated by these two outputs are constrained to have the same length.

The results reported in Tables~\ref{table:cs_parall} and~\ref{table:lv_parall} are computed with \textit{multi-bleu.perl}
which makes them consistently lower than official evaluation matrix scores\footnote{\url{http://matrix.statmt.org}}.

\subsection{Training}
\label{train_fnmt}
All models use \tbf{512}-dimensional embeddings and GRU layers with \tbf{1024} hidden units.
The validation BLEU is computed after each 20K updates and the training stops if no
improvement is achieved after 30 consecutive validations. The rest of the hyperparameters
are the same as Section~\ref{sec:nmt}.

The NMT systems are trained using all the provided bitext processed by a joint BPE model with 90K merge operations.
The sentences longer than 50 tokens are filtered out after BPE segmentation.
For FNMT systems, BPE is applied on the lemma sequence and the corresponding factors are repeated when a split occurs.

We also trained systems with synthetic data which are initialized with a previously trained model on the provided bitext only. For these systems, the learning rate is set to 0.0001 and the validations are performed every 5K updates
in order to avoid overfitting on synthetic data and forgetting the previously learned weights.
Two models with different seeds are trained for NMT and FNMT systems for ensembling purposes.



\subsection{N-best Reranking}
We experimented with different types of N-best reranking of hypotheses generated with beam search (beam size = 12) using our best FNMT. 
For each hypothesis, we generate the surface form with the factors-to-word procedure, which can be ambiguous. 
Since a single \{lemma, factors\} pair may lead to multiple possible words, $k$ possible words are considered for each pair (with $k$ being 10 for Czech and 100 for Latvian).
Finally, the hypotheses are rescored with our best word-based NMT model to select the 1-best hypothesis.

For English$\rightarrow$Latvian, we have also performed N-best reranking with two Recurrent Neural Network Language Models (RNNLM),
a simple RNNLM \cite{mikolov2010recurrent} and GRU-based RNNLM included in \tit{nmtpy}.
The RNNLMs are trained on WMT17 Latvian monolingual corpus and the target side of the available bitext (175.2M words in total).
For the FNMT system, the log probability obtained by our best word-based NMT model is also used in addition to the RNNLM scores.
The reranking is done using the \emph{nbest} tool provided by the CSLM toolkit\footnote{\url{http://github.com/hschwenk/cslm-toolkit}} \cite{schwenk10}.
(The score weights were optimized with CONDOR \cite{berghen2005condor} to maximize the BLEU score on newsdev2017 set.)

\subsection{English$\rightarrow$Czech}
The English$\rightarrow$Czech systems are trained using approximately 20M sentences from the relevant news domain parallel data provided by WMT17. 
Early stopping is performed using newstest2015 and newstest2016 is used as internal test set.
All datasets are tokenized and truecased using the Moses toolkit~\cite{Moses:2007:acl}.
PoS tagging is performed with Morphodita toolkit~\cite{strakova14morphodita} as well as the reinflection to go from factored representation to word. 
Synthetic data is generated from news-2016 monolingual corpus provided by~\citet{backtrans}.
In order to focus more on the provided bitext, five copies of news-commentary and the \emph{czeng} news dataset are added to the backtranslated data.
Also, 5M sentences from the \emph{czeng} EU corpus applying modified Moore-Lewis filtering with XenC \cite{Rousseau13}. 
We end up with about 14M sentences and 322M words for English and 292M for Czech.


\begin{table}[h] 
\begin{center}
\resizebox{1.\columnwidth}{!}{%
\begin{tabular}{lcc} 
\toprule
\textbf{System} & {\textbf{newstest2016}} & {\textbf{newstest2017}} \\ \midrule
\textbf{NMT} \\
(CS1) Baseline    					& 18.30 & 14.90 \\ 
(CS2) CS1 + synthetic     			& 24.18 & 20.26 \\ 
(CE1) Ensemble(CS2)      				& 24.52 & \tbf{20.44} \\ \midrule
\textbf{FNMT} \\
(CS3) single h2o layer  				& 17.30 & 14.19 \\ 
(CS4) sep. h2o layers   				& 17.34 & 14.73 \\ 
(CS5) CS4 + synthetic     			& 22.30 & 19.34 \\ 
(CS6) CS5 n-best reranking   		& 23.39 & 19.83 \\ %
(CE2) Ensemble(CS5) n-best reranking  & 24.05 & \tbf{20.22} \\ %
\bottomrule
\end{tabular}}
\caption{\label{table:cs_parall} EN$\rightarrow$CS. \tbf{Bold} scores represent primary submissions. Ensemble(CS$n$) correspond to the ensemble of 2 systems CS$n$ trained with different seeds.}
\end{center}
\end{table}

\subsection{English$\rightarrow$Latvian}
The English$\rightarrow$Latvian systems are trained using all the parallel data available for the WMT17 evaluation campaign.
Data selection was applied to the DCEP corpus resulting in 2M parallel sentences.
The validation set consists of 2K sentences extracted from the LETA corpus and newsdev2017 is used as internal test set.

Monolingual corpora news-2015 and 2016 were backtranslated with a Moses system~\citet{Moses:2007:acl}. 
Similarly to Czech, we added ten copies of the LETA corpus and two copies of Europarl and \emph{rapid} to perform corpus weighting.
The final corpus contains 7M sentences and 172M words for English and 143M for Latvian. 

All the Latvian preprocessing was provided by TILDE.\footnote{\url{www.tilde.com}}
Latvian PoS-tagging is done with the LU MII Tagger \cite{paikens13lv}.
Since there is no tool for Latvian to convert factors to words, all the available WMT17 monolingual data
has been automatically tagged and kept in a dictionary.
This dictionary maps the lemmas and factors to their corresponding word.
After preprocessing, we filter out training sentences with a maximum length of 50 or with a source/target length ratio higher than 3.

\begin{table}[h] 
\begin{center}
\resizebox{1.\columnwidth}{!}{%
\begin{tabular}{lcc} 
\toprule
\textbf{System} 	&  {\textbf{newsdev2017}} & {\textbf{newstest2017}} \\ \midrule
\textbf{NMT} & \\
(LS1) Baseline        	 			& 15.25 & 10.36 \\
(LS2) LS1 + synthetic     			& 21.88 & \tul{15.26} \\ 
(LS3) LS2 RNNLM reranking				& 21.98 & \tbf{15.59} \\ 
(LE1) Ensemble(LS2)  		 			& 22.34 & 15.46 \\ 
(LE2) Ensemble(LS2) RNNLM reranking	& 22.46 & 16.04 \\ \midrule
\textbf{FNMT} & \\
(LS4) single h2o layer   	& 14.45 & 10.45 \\ 
(LS5) sep. h2o layers    	& 14.39 & 10.69 \\ 
(LS6) LS5 + synthetic        	& 18.93 & \tul{13.98} \\ 
(LS7) LS6 n-best reranking 	& 21.24 & \tul{15.28} \\    
(LS8) LS6 RNNLM reranking	 	& 21.79 & \tbf{15.51} \\
(LE3) Ensemble(LS6) n-best reranking & 21.90 & 15.35 \\ 
(LE4) Ensemble(LS6) RNNLM reranking	 		& 21.87 & 15.53 \\
\bottomrule
\end{tabular}}
\caption{\label{table:lv_parall} EN$\rightarrow$LV. \tul{Underlined} and \tbf{bold} scores represent contrastive and primary submissions. Ensemble(S$n$) correspond to the ensemble of 2 systems S$n$ trained with different seeds.}
\end{center}
\end{table}


\subsection{Analysis}
We observe that including the synthetic parallel data in addition to the provided bitext results in a big improvement in NMT and FNMT for both language pairs (see systems CS2 and CS5 in Table~\ref{table:cs_parall} and LS2 and LS6 in Table~\ref{table:lv_parall}).
Applying the ensemble of several models also gives improvement for all systems (CS1-CS2 and LS1-LS4).
N-best reranking of FNMT systems (systems CS6 and LS7) shows bigger improvement when translating into Latvian than into Czech.
This is due to the quality of the dictionary used for reinflection in each language.
The Morphodita tool for Czech includes only good candidates, besides a similar tool is not available for Latvian.
The reranking with RNNLM gives an improvement for the NMT and FNMT systems when translating Latvian (LS3 and LS8).
As a follow-up work after submission, we ensembled two models applying reranking for Latvian and got improvements (LE2-LE4).
Finally, the submitted translations for NMT and FNMT systems obtain very similar automatic scores. 
However, FNMT systems explicitly model some grammatical information leading to different lexical choices, which might not be captured by the BLEU score.
Human evaluation shows for EN-LV task that NMT system obtained 43\% of standardized mean direct assessment score and FNMT system obtained 43.2\% showing a small improvement in FNMT system. Both systems obtained 55.2\% in EN-CS task.
Other analysis has been done \cite{burlot16morpheval} about morphology strength showing good results in EN-LV task.
FNMT system helps when the corpus is not huge, this is the case of EN-LV task but EN-CS dataset is huge. Therefore, NMT system has already the information to learn  the morphology.

\section{Conclusion and Discussion}
In this paper, we presented LIUM machine translation systems for WMT17 news translation task
which are among the top submissions according the official evaluation matrix.
All systems are trained using additional synthetic data which significantly
improved final translation quality.

For English\ra Turkish, we obtained (post-deadline) state-of-the-art results with a
small model ($\sim$11M params) by tying all the embeddings in the network and simplifying
the output of the recurrent decoder.
One other interesting observation is that the model trained using \tit{only} synthetic data surpassed the
one trained on genuine translation corpus. This may indicate that for low-resource
pairs, the amount of training data is much more important than the correctness
of source-side sentences.

For English\ra Czech and English\ra Latvian pairs, the best factored NMT systems performed equally well compared to NMT systems.
However, it is important to note that automatic metrics may not be suited to assess better lexical and grammatical choices made by the factored systems.

\section*{Acknowledgments}
This work was supported by the French National Research Agency (ANR) through the CHIST-ERA M2CR project, under the contract number ANR-15-CHR2-0006-01\footnote{\url{http://m2cr.univ-lemans.fr}} and also partially supported by the MAGMAT project.

\bibliography{wmt17}
\bibliographystyle{emnlp2017}

\end{document}